\documentclass[]{llncs}
\usepackage[T1]{fontenc}
\usepackage{graphicx} 
\usepackage{url}
\usepackage{hyperref}
\usepackage{caption}
\usepackage{subcaption}
\usepackage{orcidlink}

\begin{document}
\titlerunning{Comparing DT with Offline RL Algorithms}
\title{A Comparison Between Decision Transformers and Traditional Offline Reinforcement Learning Algorithms in Varying Reward Settings for Continuous Control Tasks}
\author{Ali Murtaza Caunhye\inst{1}\orcidID{0009-0002-8788-2730}\and Asad Jeewa\inst{1,2}\orcidID{0000-0003-4329-8137}}


\authorrunning{A.M. Caunhye and A. Jeewa}


\institute{School of Mathematics, Statistics and Computer Science,\\ University of KwaZulu-Natal, Durban, South Africa \and Centre for Artificial Intelligence Research (CAIR), Durban, South Africa\\ \email{221007132@stu.ukzn.ac.za, jeewaa1@ukzn.ac.za}}

\date{November 2024}

\maketitle

\pagenumbering{gobble} 
\begin{abstract}
The field of Offline Reinforcement Learning (RL) aims to derive effective policies from pre-collected datasets without active environment interaction. While traditional offline RL algorithms like Conservative Q-Learning (CQL) and Implicit Q-Learning (IQL) have shown promise, they often face challenges in balancing exploration and exploitation, especially in environments with varying reward densities. The recently proposed Decision Transformer (DT) approach, which reframes offline RL as a sequence modelling problem, has demonstrated impressive results across various benchmarks. This paper presents a comparative study evaluating the performance of DT against traditional offline RL algorithms in dense and sparse reward settings for the ANT continous control environment. Our research investigates how these algorithms perform when faced with different reward structures, examining their ability to learn effective policies and generalize across varying levels of feedback. Through empirical analysis in the ANT environment, we found that DTs showed less sensitivity to varying reward density compared to other methods and particularly excelled with medium-expert datasets in sparse reward scenarios. In contrast, traditional value-based methods like IQL showed improved performance in dense reward settings with high-quality data, while CQL offered balanced performance across different data qualities. Additionally, DTs exhibited lower variance in performance but required significantly more computational resources compared to traditional approaches. These findings suggest that sequence modelling approaches may be more suitable for scenarios with uncertain reward structures or mixed-quality data, while value-based methods remain competitive in settings with dense rewards and high-quality demonstrations.
\keywords{Offline Reinforcement Learning \and Decision Transformer \and Conservative Q-Learning \and Implicit Q-Learning \and Sequence Modeling \and ANT Environment \and Sparse Rewards \and Dense Rewards \and Continuous Control.}

\url{https://anonymous.4open.science/r/dt_vs_traditionalRL_ANT-8524/readme.md}

\end{abstract}

\section{Introduction}
Reinforcement learning algorithms face challenges in widespread adoption due to their reliance on online learning, where agents iteratively gather experience from interacting with the environment. In many scenarios, such as robotics, education, or healthcare, this process is impractical or costly\cite{prudencio_survey_2023}. For example, real-world data collection for autonomous driving is expensive, requires extensive safety measures, and carries inherent risks. In such cases, relying on previously collected data can be more practical for training agents, especially in complex environments\cite{levine_offline_2020}.
Despite these challenges, reinforcement learning has demonstrated success in domains like game playing and robotics\cite{kaelbling_reinforcement_1996}. Through trial and error, agents refine their decision-making by optimizing rewards, enabling them to learn and adapt autonomously.

Offline Reinforcement Learning (RL) has emerged as a viable alternative to online RL, particularly in scenarios where environment interaction is costly or impractical. By learning solely from a fixed dataset, offline RL eliminates the need for exploration, making it well-suited for real-world applications such as robotics and healthcare \cite{prudencio_survey_2023}. However, a core challenge of offline RL is distributional shift—when the learned policy diverges from the dataset, it may produce out-of-distribution actions that lead to poor performance or failure \cite{levine_offline_2020}. Traditional offline RL algorithms, such as Conservative Q-Learning (CQL) and Implicit Q-Learning (IQL), mitigate this issue through value function regularization and uncertainty estimation.

Recently, Decision Transformers (DT) \cite{chen_decision_2021} have emerged as an alternative to conventional value-based offline RL methods. Rather than relying on explicit value function optimization, DTs frame decision-making as a sequence modeling problem. Inspired by advancements in Natural Language Processing (NLP) \cite{vaswani_attention_2017}, DTs predict action sequences conditioned on desired returns, allowing them to leverage long-range dependencies within trajectories. This approach has demonstrated promising results, particularly in high-dimensional continuous control tasks where learning an optimal policy requires capturing intricate temporal dependencies \cite{egon_deep_2023}.

Despite these advantages, Decision Transformers have not been extensively evaluated against traditional offline RL algorithms in different reward structures. Prior studies have primarily focused on benchmarking DTs against offline RL baselines in standard datasets but have not thoroughly explored their effectiveness across varying reward densities. Reward density plays a crucial role in RL performance: dense rewards provide frequent feedback, aiding learning, whereas sparse rewards make optimization more challenging by offering limited reinforcement signals. Understanding how DTs perform relative to traditional offline RL methods in such settings remains an open question.

To address this gap, we conducted a comparative analysis of Decision Transformers and traditional offline RL algorithms (CQL, IQL) in varying reward settings within the ANT environment \cite{fu_d4rl_2020}. ANT, a quadrupedal locomotion task from the MuJoCo physics engine, presents a challenging continuous control problem with high-dimensional state-action spaces. By examining performance differences in both dense and sparse reward configurations, we provide insights into how each approach adapts to different levels of feedback frequency and quality, particularly in handling the challenges posed by sparse rewards in complex continuous control tasks.

In this study, we explored the efficacy of Decision Transformers alongside traditional offline RL algorithms, such as Conservative Q-Learning (CQL) \cite{kumar_conservative_2020} and Implicit Q-Learning (IQL) \cite{kostrikov_offline_2021}, across diverse reward structures in the ANT environment. While prior work has demonstrated the effectiveness of these methods individually, limited research has systematically compared their performance under varying reward schemes. Given that reward density significantly influences learning dynamics, this comparison was necessary to evaluate how these algorithms adapt to different levels of feedback frequency and quality. By analyzing their behavior in both dense and sparse reward settings, we aimed to identify the strengths and limitations of each approach, particularly in handling the challenges posed by sparse rewards in a complex, high-dimensional continuous control task like the ANT environment.
\section{Background}
\subsection{Offline RL Algorithms}
Offline reinforcement learning (RL) relies on pre-collected datasets rather than active environment interaction to learn policies. Two widely used algorithms for solving continuous control tasks are Conservative Q-Learning (CQL) and Implicit Q-Learning (IQL).

CQL extends Q-learning to continuous action spaces by penalizing Q-values for out-of-distribution (OOD) actions—those not present in the dataset. This regularization prevents the learned policy from overestimating unseen actions, which is a common failure mode in offline RL where environment interaction is restricted. Without such constraints, policies may drift toward unreliable state-action pairs, degrading performance \cite{kumar_conservative_2020}.

IQL, in contrast, avoids explicit Q-value regularization by strictly adhering to the dataset distribution. Instead of estimating Q-values for unseen actions, IQL extracts optimal behaviors by treating state values as stochastic variables and applying an asymmetric averaging technique. This ensures stable learning without overestimation risks, making IQL particularly effective for extracting high-quality behavior from offline datasets \cite{kostrikov_offline_2021}.

Decision Transformers (DT) offer a different perspective by reframing offline RL as a sequence modeling problem. Instead of optimizing a value function or enforcing distributional constraints, DTs predict action sequences based on desired future returns, leveraging transformer architectures originally developed for Natural Language Processing and other sequential tasks. This shift enables DTs to capture long-range dependencies in trajectory data, making them well-suited for complex continuous control tasks.

\subsection{Offline RL Datasets}
Offline RL datasets vary in quality and reward density, both of which significantly impact algorithm performance.
\begin{itemize}
    \item Dataset Quality: Refers to the average return of the trajectories in a dataset. Some datasets consist of expert demonstrations with high returns, while others contain suboptimal behaviors from weaker policies.
    \item Reward Density: Defines how frequently rewards are provided in a task. Dense reward environments offer continuous feedback (e.g., rewarding forward movement at every timestep), facilitating smooth policy updates. Sparse reward settings, on the other hand, provide feedback only upon achieving specific milestones (e.g., reaching a target), making learning more challenging.
\end{itemize}

To systematically study these factors, we used datasets from D4RL \cite{fu_d4rl_2020}, a benchmark suite for offline RL. D4RL datasets capture diverse data collection strategies, including Medium, Medium-Replay, Medium-Expert, and Expert datasets, which differ in both return quality and diversity of trajectories. Evaluating offline RL algorithms across these variations allows for a more comprehensive understanding of their adaptability to different learning conditions.

\subsection{Decision Transformer}
\begin{figure} [h!]
\centering
\includegraphics[width=0.75\linewidth]{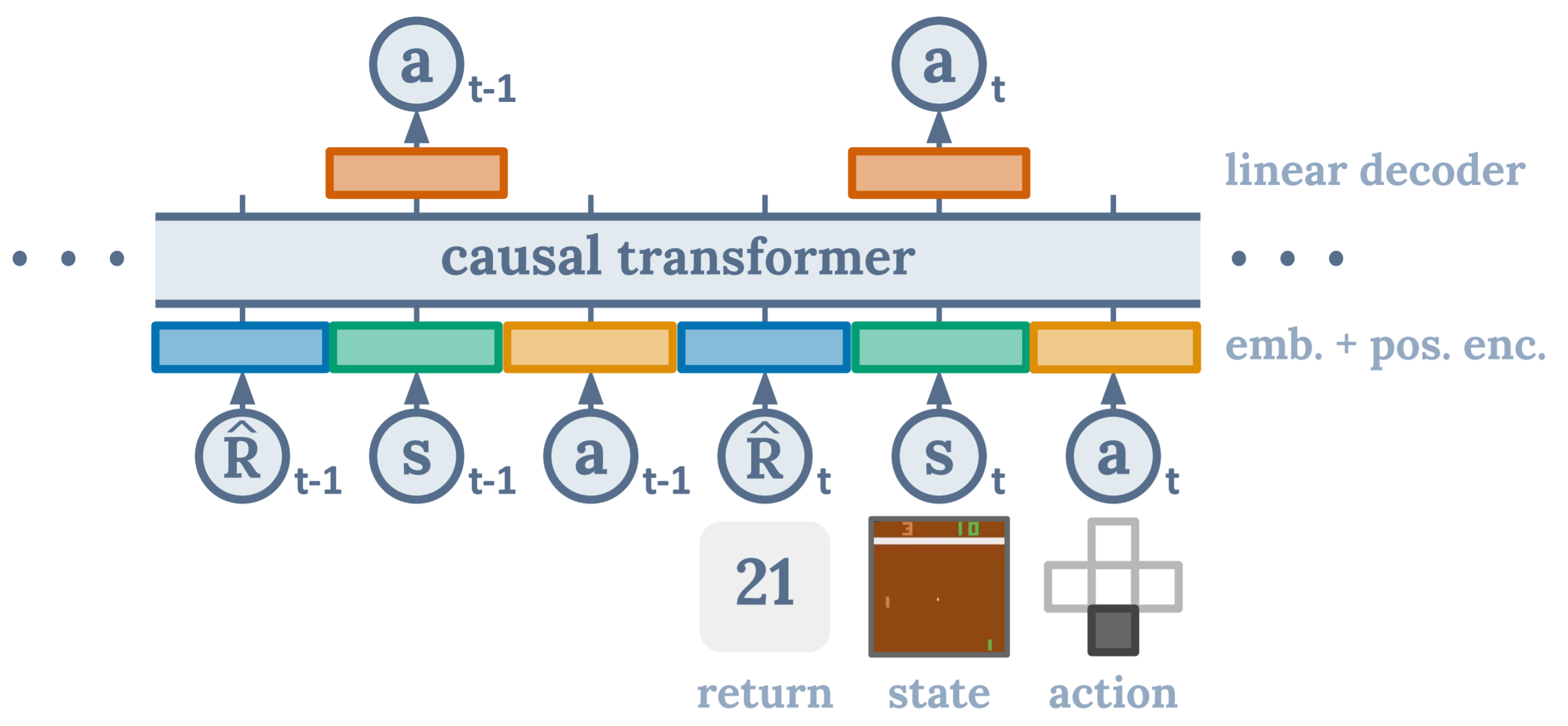}
\caption{Decision Transformer\cite{chen_decision_2021}}
\label{decision_transformer}
\end{figure}

Decision Transformers (DT) are a specific type of transformer architecture adapted for RL tasks\cite{chen_decision_2021} (shown in~\autoref{decision_transformer}. The main idea behind Decision Transformers is to cast the RL problem as a conditional sequence modelling problem, where the goal is to generate a sequence of actions that achieve a desired return (cumulative reward). This differs from traditional RL approaches, which typically involve fitting a value function or computing policy gradients\cite{kumar_conservative_2020}. Decision Transformers use a transformer architecture to model the joint distribution of states, actions, and rewards and can be trained offline using collected trajectory data.

Takyar \cite{takyar_decision_2023} outlined the core components of a Decision Transformer (DT). Built on a transformer architecture, DT leverages multi-headed self-attention to model dependencies in sequential data. It learns from offline State-Action-Reward (SAR) sequences, conditioning its predictions on past experiences to generate optimal action sequences. By modeling rewards directly, DT predicts actions that maximize cumulative returns without traditional value function optimization. This approach enables DT to effectively learn policies in offline RL settings by capturing long-term dependencies in historical data. 

DTs are particularly useful in offline settings because they learn directly from fixed datasets without requiring environment interaction, making them effective in environments where exploration is costly or unsafe. For example, in self-driving cars, where testing unsafe actions in the real world would be highly risky. They can also leverage large sequences of past experiences to model long-term dependencies, 

Transformers revolutionalized the field of NLP, and recent studies have shown promising results of applying the transformer architecture to fields such as Computer Vision\cite{jamil_comprehensive_2023} and Audio, Multi-modal Processing\cite{akbari_vatt_2021} and Reinforcement Learning\cite{chen_decision_2021}. 

In this paper, we present a comparative study between decision transformers and traditional offline RL algorithms (CQL and IQL) across different continuous control task environments to garner attention and further research into integrating transformers into the field of reinforcement learning.
\section{Literature Review}
This work explores three key areas related to RL in continuous control tasks: deep RL approaches, offline RL methods, and the application of transformers in reinforcement learning for continuous control tasks.

\subsection{Offline RL in Continuous Control Tasks}
Kumar et al.\cite{kumar_conservative_2020} introduced the Conservative Q-Learning (CQL) framework to address the limitations of standard off-policy RL methods in offline settings. CQL's approach of learning a conservative Q-function that lower-bounds the true policy value proved particularly relevant for our study, as it provides theoretical guarantees on performance across different reward structures. The authors demonstrated that CQL significantly outperformed existing offline RL methods on both discrete and continuous control domains, achieving final returns 2-5 times higher, especially when handling complex and multi-modal data distributions like those we examine in our work.

Building on this work, Kostrikov et al. \cite{kostrikov_offline_2021} introduced Implicit Q-Learning (IQL), an offline RL algorithm that took a different approach to addressing the overestimation bias in Q-learning. IQL enabled effective learning without direct Q-function maximization by employing an implicit approach to policy improvement. This characteristic makes it particularly interesting for our comparison across dense and sparse reward settings, as it potentially offers different trade-offs in handling varying reward signals. The authors demonstrated IQL's effectiveness by setting new benchmarks on the D4RL dataset \cite{fu_d4rl_2020}, which we also use in our evaluation.

While this study compares CQL and IQL with Decision Transformers, other promising approaches exist in the offline RL space. Notably, Fujimoto et al.\cite{fujimoto_off-policy_2019} proposed BCQ (Batch-Constrained deep Q-learning), which uses a generative model to approximate the action distribution of the dataset. The investigation of BCQ's performance across different reward densities represents an interesting direction for future work.

\subsection{Transformers in RL for Continuous Control Tasks}
Chen et al.\cite{chen_decision_2021} introduced the Decision Transformer, applying the transformer architecture to reinforcement learning tasks. They evaluated their approach on the D4RL benchmark\cite{fu_d4rl_2020}, which includes an array of continuous control tasks. While their study showed that Decision Transformer demonstrated competitive or superior performance compared to state-of-the-art model-free offline RL algorithms like CQL, particularly in long-horizon tasks and those requiring complex reasoning, they did not specifically analyze the ANT environment's unique challenges. The ANT environment's complex quadrupedal locomotion and high-dimensional action space presents a distinct challenge for policy learning that warrants dedicated investigation, particularly across different reward structures and dataset qualities. 

Our study addresses this gap by specifically examining how DT compares to CQL and IQL in the ANT environment under varying reward densities and data quality conditions. A limitation noted in the original DT work was that the approach struggled with very long sequences due to the quadratic computational complexity of self-attention, making the ANT environment's extended action sequences particularly interesting to evaluate.

Janner et al.\cite{janner_offline_2021} expanded on this work with their Trajectory Transformer, which they tested on similar continuous control environments in D4RL. Their approach showed strong performance in offline RL settings and exhibited impressive zero-shot generalization to new tasks. The Trajectory Transformer excelled in environments requiring long-term planning, such as AntMaze.

Several variations of Decision Transformers (DT) have been proposed to enhance their performance in offline reinforcement learning. These include Critic-Guided Decision Transformer (CGDT) \cite{wang_critic-guided_2024}, which integrates value estimation, Elastic Decision Transformer\cite{wu2023elastic}, which adapts to varying sequence lengths, and Online Decision Transformer, which extends DTs to online settings. While these advancements address specific limitations, the focus of this work was to analyze the fundamental Decision Transformer model and its limitations, particularly in handling varying reward structures and dataset qualities in continuous control tasks.

While the applicability of Decision Transformers was discussed extensively in~\cite{bhargava_when_2023}, several open challenges remain in applying offline reinforcement learning to continuous control tasks: 
\begin{enumerate}
    \item The computational complexity of transformer-based approaches with long sequences.
    \item The need for more comprehensive comparisons across different reward structures and dataset qualities and the lack of understanding of how these approaches perform under varying data quality and reward sparsity levels.
\end{enumerate} 
Addressing these challenges is crucial for advancing the effectiveness of offline RL methods in real-world applications.
\section{Methodology}
Our study utilizes the ANT environment from the MuJoCo physics engine, which is part of the D4RL benchmark\cite{fu_d4rl_2020}. This environment involves controlling a four-legged robot to walk efficiently, presenting a complex continuous control task with a high-dimensional action space.
\begin{figure}[h!]
\centering
\includegraphics[width=0.2\linewidth]{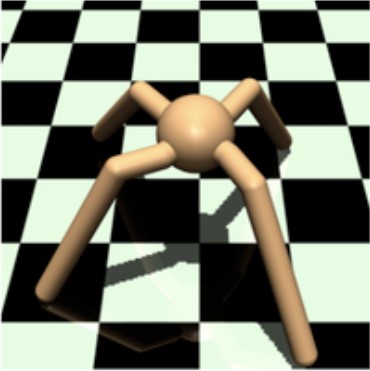}
\caption{ANT Environment\cite{fu_d4rl_2020}}
\label{ant_env}
\end{figure}
We investigate the performance of Decision Transformers (DT), Conservative Q-Learning (CQL), and Implicit Q-Learning (IQL) in both dense and sparse reward settings within the ANT environment.

\subsection{Dense and Sparse Rewards}
\begin{enumerate}
    \item Dense Reward: The agent receives feedback at each time step using the default reward structure as provided by D4RL\cite{fu_d4rl_2020}.
    \item Sparse Reward: We set the 75-percentile of all returns (the sum of rewards) in the dataset as return threshold. We assigned a reward of 0 to all trajectories in the lower 75\% of returns and a reward of 1 to the top 25\% of trajectories if the cumulative dense reward surpasses the threshold, which means the agent reaches the "goal", following the approach of Cen et al. \cite{cen_learning_2023}.
\end{enumerate}

\subsection{Dataset Types}
We utilized four distinct dataset types from D4RL, each reflecting different data collection strategies and policy performance levels. The Medium dataset consisted of trajectories generated by a medium-performing policy, where the agent received rewards around the mid-range of 4000 in the ANT environment. The Medium-Replay dataset included a more diverse set of experiences, combining data from both the medium policy and the training process leading up to it, capturing a broader distribution of state-action pairs. The Medium-Expert dataset blended medium-quality trajectories with expert demonstrations, incorporating both intermediate and high-performance behavior. Finally, the Expert dataset was composed of data collected from an expert policy, where the agent consistently achieved rewards closer to the upper range of 6000. These datasets enabled us to systematically assess how each algorithm performed under varying levels of data quality and diversity, providing valuable insights into their adaptability and generalization capabilities in offline reinforcement learning.

\subsubsection{Model Architectures}
Following prior work\cite{chen_decision_2021,kumar_conservative_2020,kostrikov_offline_2021}, we implement three different models(DT, CQL and IQL). The key parameters for each model can be found in Appendix A. 




\subsection{Experimental Setup}
We tested DT, CQL, and IQL models using CORL implementation\cite{tarasov2022corl} of offline RL models. We conducted experiments for each algorithm across all four dataset types in dense and sparse reward settings. Each experiment was run with 4 different random seeds, and the results were averaged to ensure statistical significance. We found that the hyperparameters from the Conservative Density Estimation\cite{cen_learning_2023} for CQL and IQL, and Chen et al.\cite{chen_decision_2021} were best suited for gym environments. The best results from our random search performed worse and were much more inconsistent than those presented in Cen et al.\cite{cen_learning_2023}

\subsection{Evaluation Metrics}
Our evaluation framework employs four key metrics to assess algorithm performance. The primary metric is the D4RL Normalized Score. This standardized measure scales performance between 0 and 100, where 0 represents random policy performance, and 100 represents expert policy performance, enabling consistent comparisons across different tasks and datasets. We also track the Maximum Return, which captures the highest score achieved across evaluation episodes, providing insight into the peak performance capabilities of each algorithm. We measure Return Variance by analyzing performance variability across multiple random seeds to assess reliability. Finally, we evaluate computational efficiency through Training Time, measured as the duration required to complete 100,000 timesteps across four random seeds, providing practical insights into implementation costs.


\section{Results and Discussion}
While prior work has established the effectiveness of these algorithms across various D4RL environments, our work extends these analyses by systematically comparing all three algorithms in both sparse and dense reward structures within this challenging continuous control task.

The normalized scores reported in Table \ref{tab:results} include standard deviations representing one standard deviation of uncertainty calculated across four random seeds. Notably, several configurations achieve scores exceeding 100, which represents expert-level performance. This phenomenon, also observed in prior work by \cite{kumar_conservative_2020} and \cite{kostrikov_offline_2021}, occurs when algorithms discover optimizations that surpass the original expert demonstrations, often by finding more efficient movement patterns or novel solutions not present in the training data.

\begin{table*}[h]
\caption{Normalized Performance Scores (Mean ± Standard Deviation) of DT, CQL, and IQL Across ANT Environment Datasets}
\centering
\resizebox{1.1\textwidth}{!}{%
\begin{tabular}{|c|ccc|ccc|}
\hline
& \multicolumn{3}{c|}{Sparse Dataset} & \multicolumn{3}{c|}{Dense Dataset} \\
Dataset & DT & CQL & IQL & DT & CQL & IQL \\
\hline
ant-medium-v2 & 87.9 $\pm$ 3.4 & \textbf{91.55} $\pm$ 8.56 & 84.49 $\pm$ 11.38 & 88.0 $\pm$ 5.2 & \textbf{99.49} $\pm$ 5.96 & 95.5 $\pm$ 8.6 \\
ant-medium-replay-v2 & 66.3 $\pm$ 9.7 & \textbf{71.99} $\pm$ 11.13 & 42.14 $\pm$ 7.07 & 88.07 $\pm$ 3.61 & 92.99 $\pm$ 8.81 & \textbf{97.5} $\pm$ 5.0 \\
ant-medium-expert-v2 & \textbf{120.6} $\pm$ 1.1 & 103.38 $\pm$ 15.51 & 85.95 $\pm$ 21.18 & 90.24 $\pm$ 3.39 & 107.0 $\pm$ 21.2 & \textbf{124.2} $\pm$ 5.8 \\
ant-expert-v2 & 119.3 $\pm$ 2.9 & 120.51 $\pm$ 6.42 & \textbf{120.62} $\pm$ 14.12 & 122.52 $\pm$ 3.67 & 122.0 $\pm$ 16.4 & \textbf{125.2} $\pm$ 10.1 \\
\hline
\end{tabular}%
}
\label{tab:results}
\end{table*}


\subsection{Performance Analysis Across Dense and Sparse Reward Data Structures}

\subsubsection{Sparse Reward Settings}
\begin{figure}[h]
\centering
\begin{subfigure}[t]{\textwidth}
    \centering
    \includegraphics[width=1\linewidth]{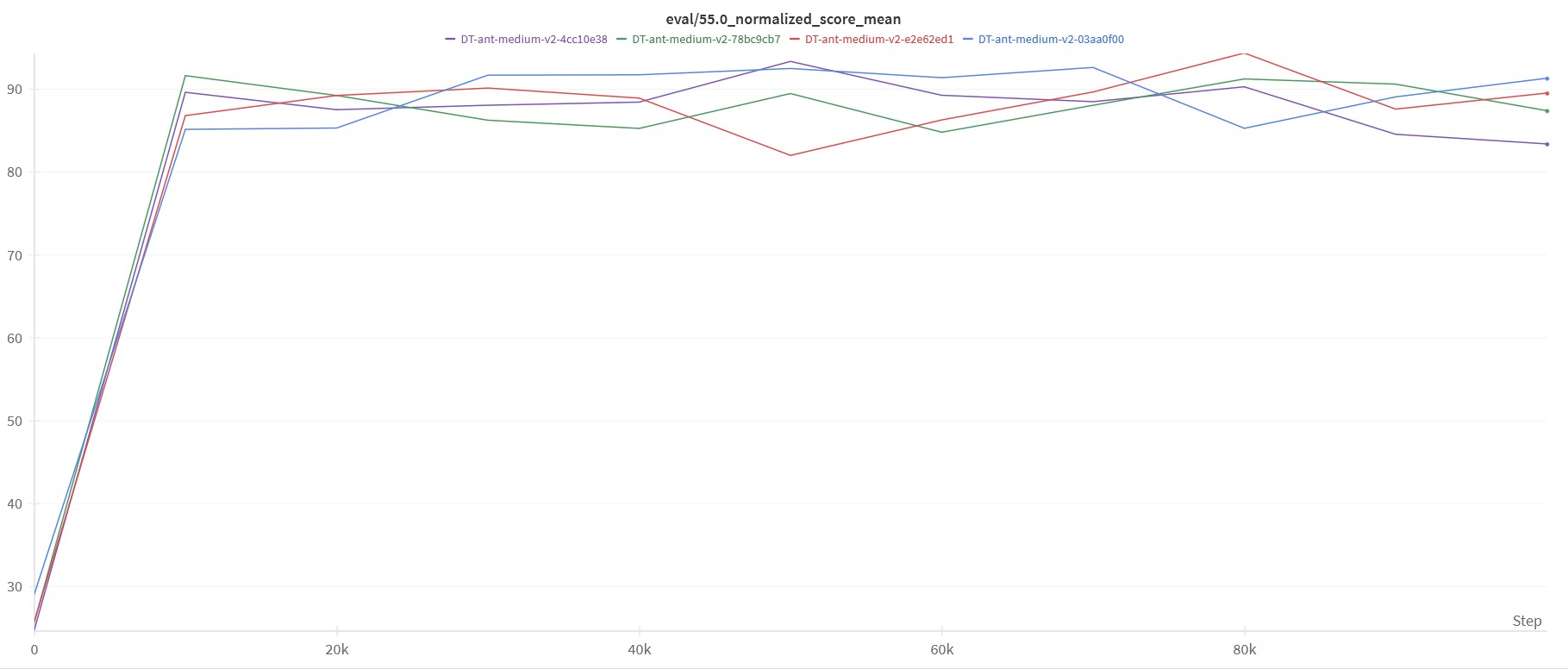}
    \caption{DT}
    \label{fig:dt_sparse_medium}
\end{subfigure}
\begin{subfigure}[t]{\textwidth}
    \centering
    \includegraphics[width=1\linewidth]{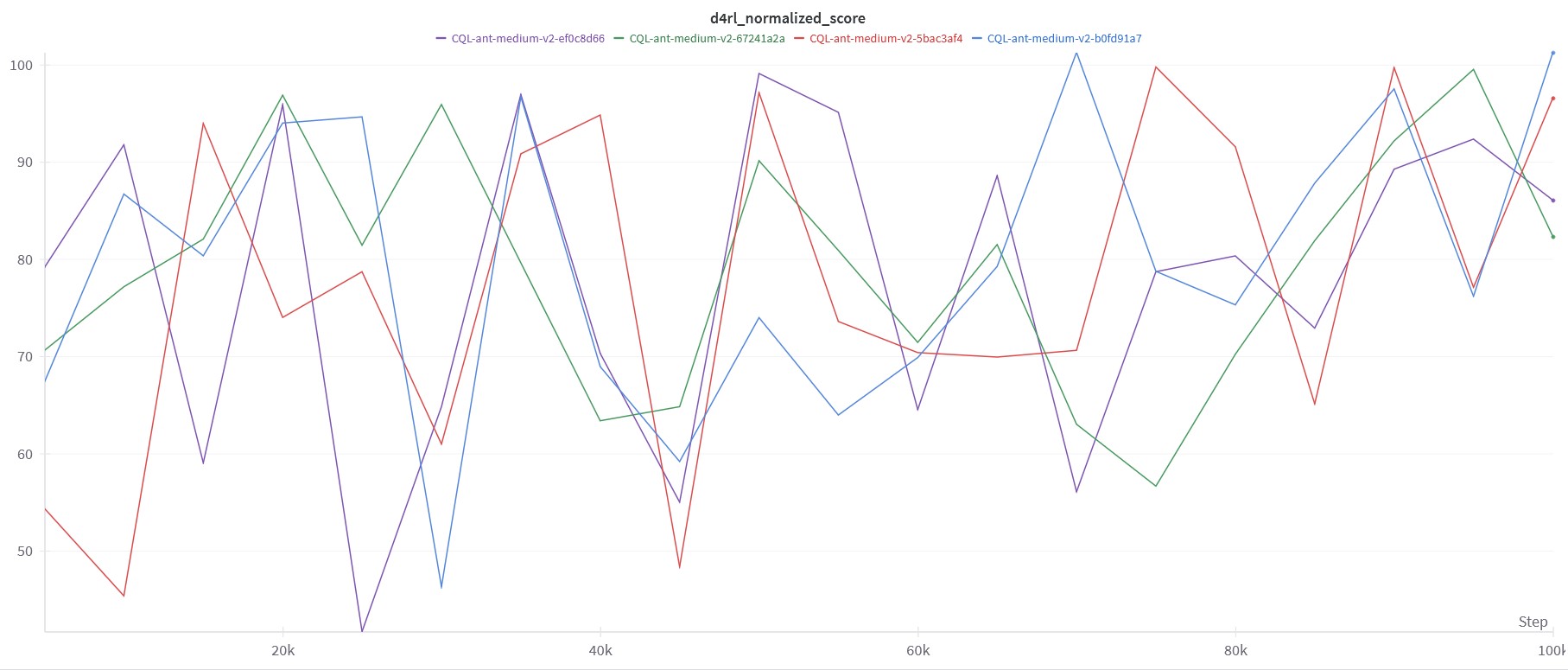}
    \caption{CQL}
    \label{fig:cql_sparse_medium}
\end{subfigure}
\begin{subfigure}[t]{\textwidth}
    \centering
    \includegraphics[width=1\linewidth]{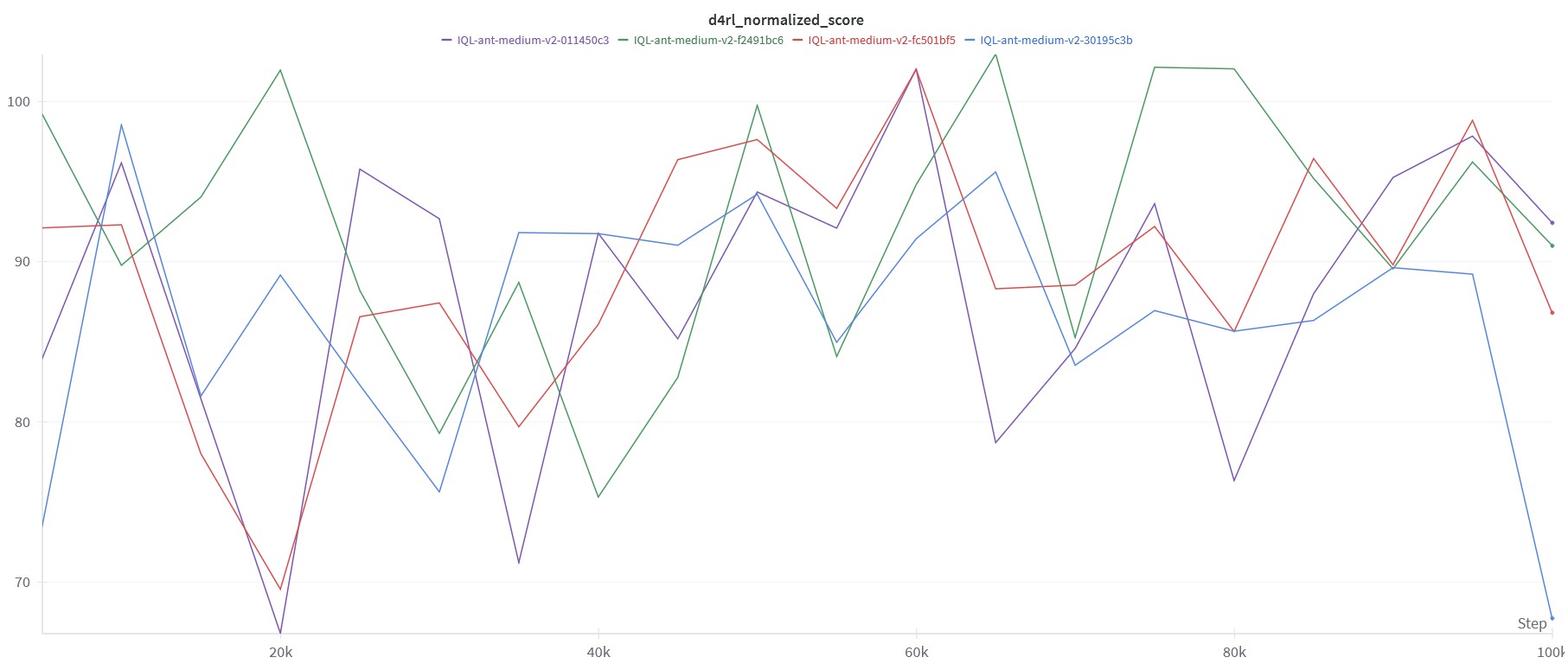}
    \caption{IQL}
    \label{fig:iql_sparse_medium}
\end{subfigure}
    \caption{D4RL normalized score graph of 4 random seeds over 100000 timesteps on ANT medium Sparse Dataset.}
    \label{fig:all_sparse_medium}
\end{figure}
From Table \ref{tab:results}, we observe varying performances across different dataset types in sparse reward scenarios. For the ant-medium-v2 dataset, CQL demonstrates superior performance, followed closely by DT, while IQL shows the lowest performance. This suggests CQL's effectiveness in extracting useful information from medium-quality data, though DT's lower variance indicates more stable performance. 

In the ant-medium-replay-v2 dataset, CQL maintains its good performance, followed by DT, with IQL significantly underperforming, indicating both CQL and DT's superior ability to handle diverse data under sparse rewards for the ANT environment.

The ant-medium-expert-v2 dataset reveals an interesting shift, with DT significantly outperforming both CQL and IQL (120.6 ± 1.1 vs 103.38 ± 15.51 and 85.95 ± 21.18 respectively), demonstrating DT's particular aptitude for leveraging mixed medium-expert data in sparse reward settings. 

For the ant-expert-v2 dataset, all three algorithms achieve comparable high performance with scores above 119, suggesting that high-quality expert data enables near-optimal performance even in sparse reward scenarios.

Figure \ref{fig:all_sparse_medium} demonstrates that the Decision Transformer exhibits the most stable learning trajectory, maintaining consistent performance between 80-90 normalized score across all four random seeds on the ANT medium with minimal fluctuations, particularly in the latter half of training. CQL shows more variation between seeds and larger performance fluctuations, occasionally achieving higher peak performances exceeding 95 on the normalized score, but with more pronounced oscillations suggesting less stable learning dynamics compared to DT. IQL demonstrates the highest variance among the three algorithms, showing significant fluctuations both within individual seeds and across different seeds, with performance occasionally dropping as low as 70 on the normalized score during training, indicating greater sensitivity to the sparse reward structure and potentially requiring more careful hyperparameter tuning.

\subsubsection{Dense Reward Settings}
\begin{figure}[h]
\centering
\begin{subfigure}[t]{\textwidth}
    \centering
    \includegraphics[width=1\linewidth]{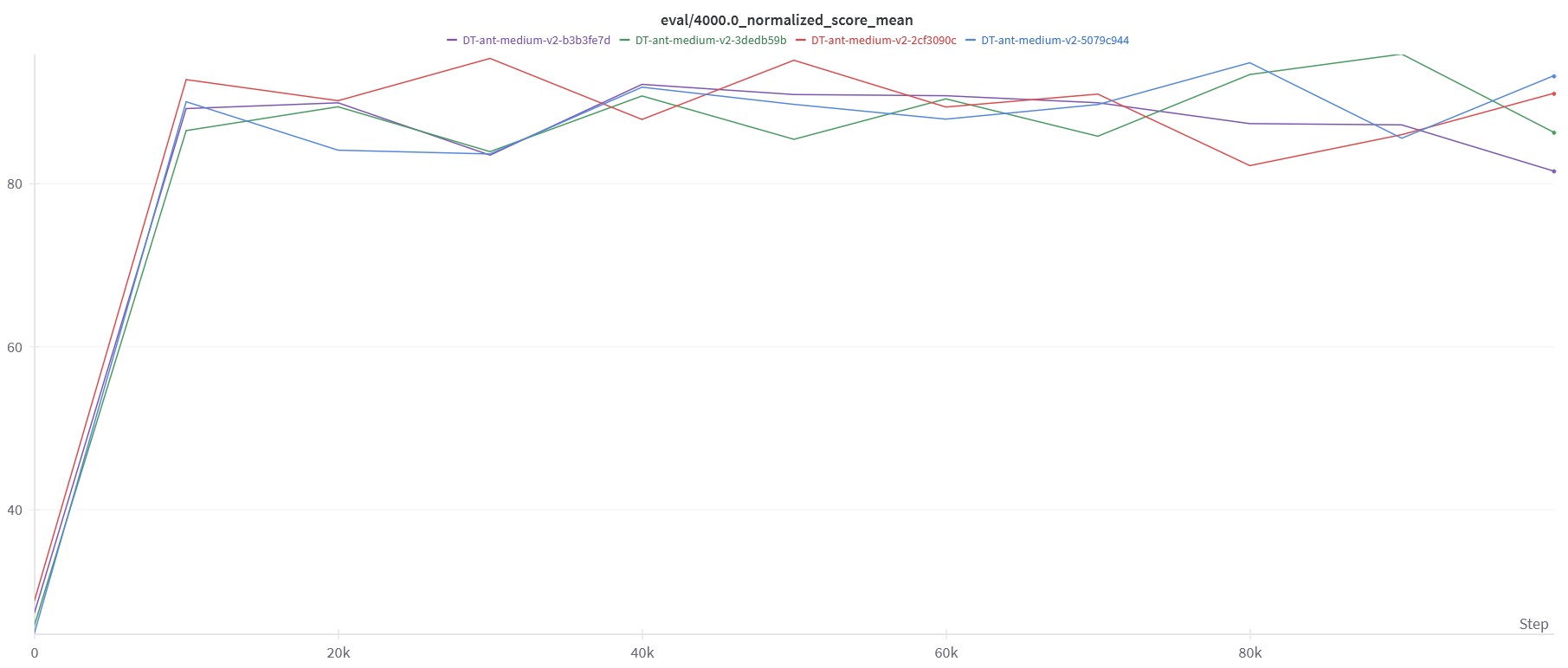}
    \caption{DT}
    \label{fig:dt_dense_medium}
\end{subfigure}
\begin{subfigure}[t]{\textwidth}
    \centering
    \includegraphics[width=1\linewidth]{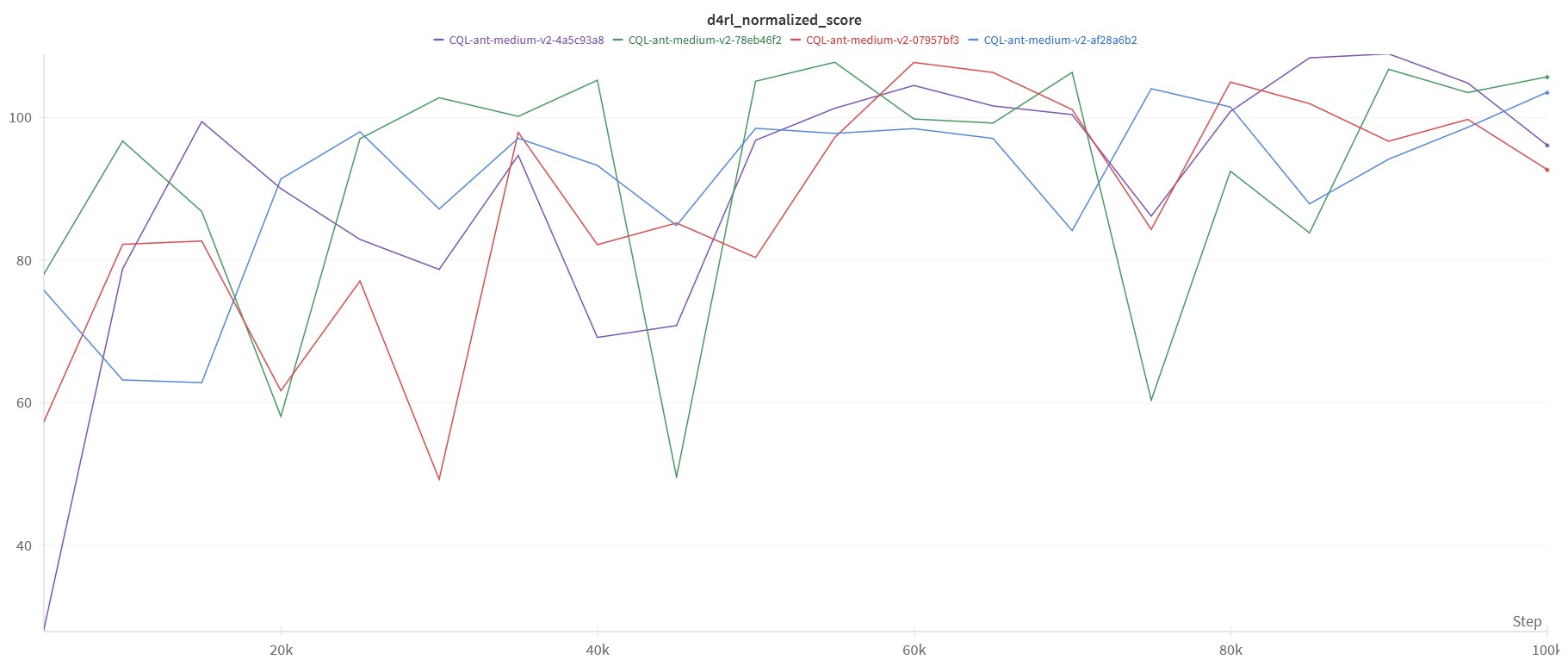}
    \caption{CQL}
    \label{fig:cql_dense_medium}
\end{subfigure}
\begin{subfigure}[t]{\textwidth}
    \centering
    \includegraphics[width=1\linewidth]{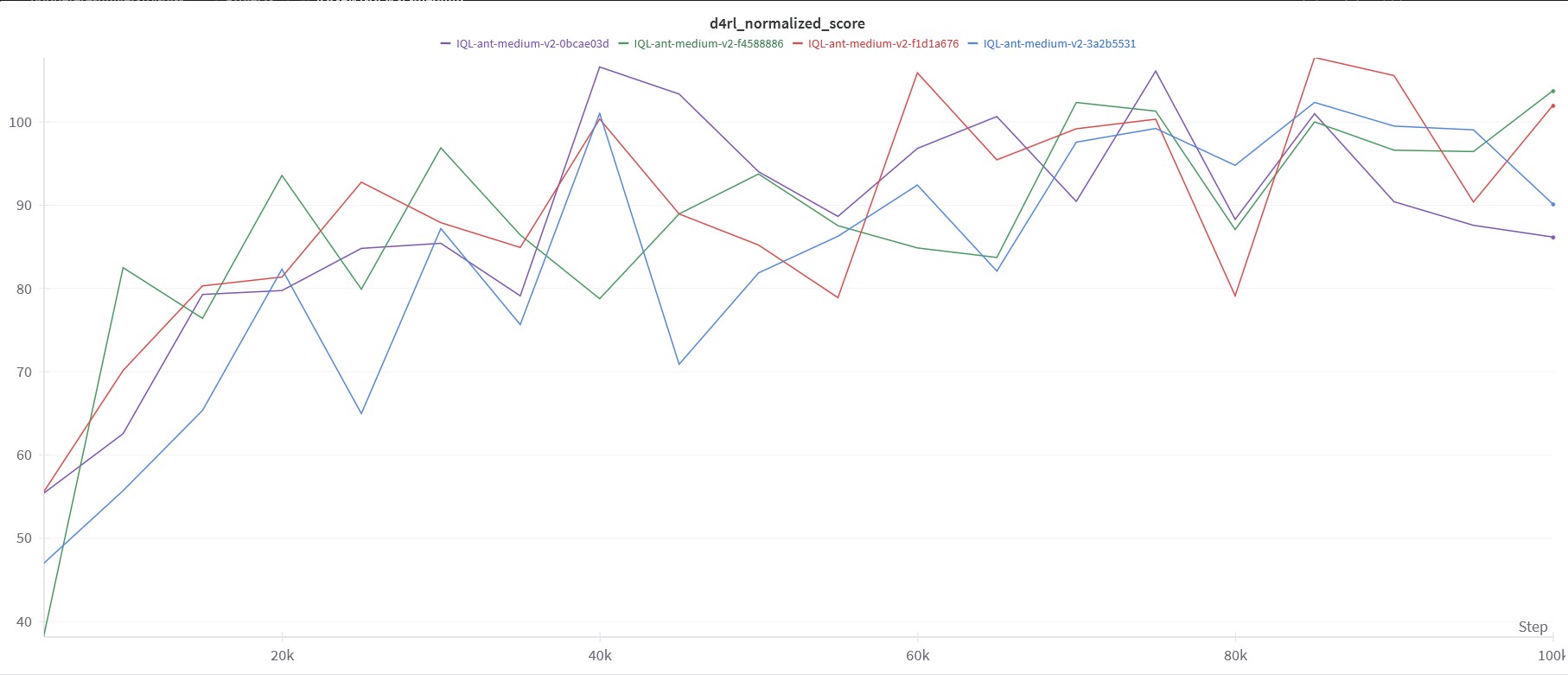}
    \caption{IQL}
    \label{fig:iql_dense_medium}
\end{subfigure}
\caption{Normalized score of 4 random seeds over 100000 timesteps on ANT medium Dense Dataset.}
    \label{fig:all_dense_medium}
    \end{figure}
The dense reward setting presents notably different performance patterns. In the ant-medium-v2 dataset, CQL performs the best (99.49 ± 5.96), followed closely by IQL (95.5 ± 8.6), with DT slightly behind (88.0 ± 5.2), indicating traditional RL methods' advantage in dense reward settings with medium-quality data. The ant-medium-replay-v2 dataset shows IQL outperforming both CQL and DT (97.5 ± 5.0 vs 92.99 ± 8.81 and 88.07 ± 3.61), highlighting IQL's effectiveness with diverse data in dense reward scenarios.

In the ant-medium-expert-v2 dataset, IQL demonstrates significant superiority over both CQL and DT (124.2 ± 5.8 vs 107.0 ± 21.2 and 90.24 ± 3.39), marking a notable reversal from the sparse reward setting and suggesting IQL's particular strength in leveraging mixed medium-expert data with dense rewards. The ant-expert-v2 dataset shows exceptional performance across all three algorithms, with scores above 122, confirming that high-quality expert data combined with dense rewards enables optimal or near-optimal performance in the ANT environment.

As shown in Figure~\ref{fig:all_dense_medium}, DT exhibits extremely low variance from early in the training process, quickly converging to a normalized score around 85-90 and maintaining this performance with minimal fluctuations across all four random seeds. This illustrates the stability of the model. In contrast, IQL demonstrates more aggressive learning with higher peak performance, reaching normalized scores above 100, but shows greater variability both within and across seeds throughout the training process. CQL displays an interesting middle ground, achieving higher peak performances than DT (occasionally exceeding 105 on the normalized score) while maintaining better stability than IQL, though still showing more variance than DT. However, CQL appears to be more susceptible to performance drops, as evidenced by occasional dips to around 60 on the normalized score during training. 

While dense rewards generally improved performance across algorithms in our study, it's crucial to note that denser reward structures are not universally superior in reinforcement learning tasks. Reward engineering - the process of designing and tuning reward functions - remains one of the most challenging aspects of RL, as over-engineered reward functions can lead to unintended behaviours, reward exploitation, or brittle policies that fail to generalize\cite{cen_learning_2023}.

Building upon these observations from the ANT medium dataset analysis, similar training dynamics and performance patterns were observed across other dataset variations (medium-replay, medium-expert, and expert), with the key difference being generally improved performance metrics as the dataset quality increased. Notably, while the relative characteristics of stability and variance between DT, CQL, and IQL remained consistent across dataset types, the baseline performance elevated progressively from medium to expert datasets, with expert datasets showing the highest normalized scores and most stable learning curves across all three algorithms. For all results of D4RL normalized score on each dataset with each algorithm, please refer to the supplementary material.

\subsection{Comparative Analysis}

\subsubsection{Dense vs Sparse Rewards}
Our analysis reveals that DT maintains lower variance across both reward structures, exhibiting relatively stable scores between sparse and dense settings. This consistency likely stems from the transformer architecture's inherent ability to model long-term dependencies, as described in Chen et al.\cite{chen_decision_2021}. Traditional RL methods (CQL and IQL) demonstrate greater reward density sensitivity, which is particularly evident in their improved performance with dense rewards in medium-quality datasets.

\subsubsection{Dataset Quality Effects}
Though their scaling characteristics differ, all algorithms exhibit performance improvements with increasing dataset quality. DT's superior performance with medium-expert data in sparse settings (120.6 ± 1.1) suggests particular effectiveness in leveraging mixed-quality data when rewards are infrequent. This may be attributed to the transformer's sequence modelling capability, which allows it to connect sparse rewards better with their generating action sequences.

\subsubsection{Computational Requirements}
Table \ref{tab:training-times} presents the computational requirements for each algorithm. While DT demonstrates strong performance characteristics, its average training time over all runs(7.5 hours) represents a significant computational overhead compared to CQL (5 hours) and IQL (2 hours). This increased computational cost stems from the transformer architecture's self-attention mechanisms and the processing of full trajectories as sequences.

\begin{table*}[h]
\caption{Training time comparison across algorithms}
\centering
\begin{tabular}{|c|c|}
\hline
Algorithm & Training Duration (hours)\\ 
\hline
DT & 7.5 \\
CQL & 5.0 \\
IQL & 2.0 \\
\hline
\end{tabular}
\label{tab:training-times}
\end{table*}

\subsubsection{Limitations and Generalizability}
While our results provide valuable insights into the relative strengths of these algorithms, it's important to note that these findings are specific to the ANT environment. Generalization to other continuous control tasks or different domains would require additional investigation. The performance characteristics we observe may be influenced by the specific dynamics and complexity of the ANT environment, and different patterns might emerge in other contexts.

These results suggest important considerations for practitioners selecting offline RL algorithms. The choice between DT, CQL, and IQL should consider not only performance metrics but also computational constraints, dataset quality, and reward structure characteristics. Further research across additional environments and reward structures would be valuable in establishing the generalizability of these findings.

\section{Conclusion}
This study presents a systematic comparison of Decision Transformers, Conservative Q-Learning, and Implicit Q-Learning in the ANT environment, examining their performance across sparse and dense reward structures with varying dataset qualities. Our investigation reveals that the algorithms exhibit distinctive strengths in different scenarios: DT demonstrates consistent performance across reward structures with notably low variance, particularly excelling in sparse reward settings with mixed-quality data (medium-expert). CQL provides balanced performance across different scenarios, while IQL shows superior performance in dense reward settings, especially with high-quality data, albeit with higher variance.

Amongst our findings is the trade-off between computational requirements and performance characteristics. While DT offers robust performance across different reward structures, it requires significantly more computational resources compared to traditional approaches. This presents an important consideration for practical applications, where resource constraints may influence algorithm selection.

We also observe that exceeding expert-level performance (scores above 100) is achievable across all three algorithms in certain configurations, particularly with higher-quality datasets. This finding aligns with previous work\cite{kumar_conservative_2020}\cite{kostrikov_offline_2021} in offline RL, where algorithms can discover optimizations beyond the original expert demonstrations.

These insights, while specific to the ANT environment, provide valuable guidance for the offline RL community.

This work aligns with the conference theme of Sustainable Digital Intelligent Frontiers by exploring the use of advanced AI algorithms, such as Decision Transformers, in solving complex continuous control tasks. The study's emphasis on offline reinforcement learning (RL) addresses sub-themes such as AI and digital technologies for sustainable development by enhancing the efficiency and effectiveness of RL algorithms, potentially reducing the need for costly online interactions and making these algorithms more accessible for real-world applications.

\section{Possible Extensions}
While our study provides valuable insights into the performance characteristics of Decision Transformers and traditional offline RL algorithms in the ANT environment, several promising directions for future research emerge. A primary extension would be to scale these experiments across additional environments and benchmarks to verify whether the observed patterns generalize beyond the ANT environment. This could include other MuJoCo environments, different locomotion tasks, and manipulation scenarios, providing a more comprehensive understanding of each algorithm's strengths and limitations across varied contexts.

The real-world applicability of these findings could be explored through studies involving real-world datasets, to assess how these algorithms handle constraints and environmental uncertainties.

Further research could also investigate variations in transformer architectures, such as different attention mechanisms and layer configurations, for improved performance and efficiency. Additionally, exploring online fine-tuning of offline-trained models could provide insights into the potential for better sample efficiency and performance improvements.

Finally, expanding the study to include a broader range of datasets with varying qualities and characteristics would provide deeper insights into how these algorithms handle different data distributions and reward structures, helping guide future algorithm development in offline reinforcement learning.

\bibliographystyle{splncs04}

\end{document}